\documentclass[runningheads]{lncs} 

\usepackage{graphics} % for pdf, bitmapped graphics files
\usepackage{epsfig} % for postscript graphics files
\usepackage{times} % assumes new font selection scheme installed
\usepackage{amsmath} % assumes amsmath package installed
\usepackage{amssymb}  % assumes amsmath package installed
\usepackage{comment}
\usepackage{xspace}
\usepackage{xcolor}
\usepackage{booktabs}
\usepackage{pifont}
\usepackage{algorithm}
\usepackage{algpseudocode}
\usepackage{cleveref}

\newcommand{\dynsys}{\ensuremath{\mathcal{N}}\xspace}
\newcommand{\poly}{\ensuremath{\mathcal{P}}\xspace}
\newcommand{\simp}{\ensuremath{S}\xspace}
\newcommand{\pset}{\ensuremath{P}\xspace}
\newcommand{\ints}{\ensuremath{\mathbb{Z}}\xspace}
\newcommand{\nats}{\ensuremath{\mathbb{N}}\xspace}
\newcommand{\reals}{\ensuremath{\mathbb{R}}\xspace}
\newcommand{\conv}{\ensuremath{\mathit{conv}}\xspace}
\newcommand{\verts}{\ensuremath{\mathit{vert}}\xspace}

\newcommand{\ftraj}{\ensuremath{\mathbf{F}}\xspace}
\newcommand{\btraj}{\ensuremath{\mathbf{B}}\xspace}
\newcommand{\nextt}{\ensuremath{\mathit{next}}\xspace}

\newcommand{\prevg}{\ensuremath{\mathit{prev}_G}\xspace}
\newcommand{\preva}{\ensuremath{\mathit{prev}_A}\xspace}
\newcommand{\noise}{\ensuremath{E}\xspace}
\newcommand{\model}{\ensuremath{\mathcal{M}}\xspace}

\newcommand{\init}{\ensuremath{I}\xspace}
\newcommand{\dom}{\ensuremath{D}\xspace}
\newcommand{\trans}{\ensuremath{\vec{F}}\xspace}
\newcommand{\ctrl}{\ensuremath{\vec{u}}\xspace}
\newcommand{\timestep}{\ensuremath{\delta}\xspace}
\newcommand{\horizon}{\ensuremath{T}\xspace}
\newcommand{\reach}{\ensuremath{G}\xspace}
\newcommand{\avoid}{\ensuremath{A}\xspace}
\newcommand{\state}{\ensuremath{\vec{x}}\xspace}

\newcommand{\stateset}{\ensuremath{\mathcal{X}}\xspace}
\newcommand{\stateBig}{\stateset}

\newcommand{\powset}[1]{\ensuremath{2^{#1}}\xspace}
\newcommand{\vleq}{\ensuremath{\preccurlyeq}\xspace}
\newcommand{\vgeq}{\ensuremath{\succcurlyeq}\xspace}
\newcommand{\unit}[1]{\ensuremath{\mathbf{e}_{#1}}\xspace}

\newcommand{\tora}{\ensuremath{\mathcal{T}}\xspace}
\newcommand{\uni}{\ensuremath{\mathcal{U}}\xspace}
\newcommand{\att}{\ensuremath{\mathcal{R}}\xspace}
\newcommand{\fab}{\textsc{FaBRIC}\xspace}
\newcommand{\ssharp}{\textsc{SHARP}\xspace}
\newcommand{\clean}{\textsc{CLEAN}\xspace}
\newcommand{\crisp}{\textsc{CRISP}\xspace}
\newcommand{\dripy}{\textsc{DRiPy}\xspace}
\newif\ifcomments 
\commentstrue
\ifcomments

\makeatletter
\newcommand{\noindentafterdisplay}{%
  \@afterindentfalse\@afterheading}
\makeatother

\title{\LARGE \bf
The \fab Strategy for Verifying Neural Feedback Systems
}

\titlerunning{The \fab Strategy}

\author{Samuel I. Akinwande\inst{1}, Sydney M. Katz\inst{1}, Mykel J. Kochenderfer\inst{1}, and Clark Barrett\inst{2}% <-this 
}
\institute{
Department of Aeronautics and Astronautics, Stanford University, Stanford, USA \and
Department of Computer Science, Stanford University, Stanford, USA
}

\authorrunning{Akinwande et al.}

\begin{document}
\maketitle

%%%%%%%%%%%%%%%%%%%%%%%%%%%%%%%%%%%%%%%%%%%%%%%%%%%%%%%%%%%%%%%%%%%%%%%%%%%%%%%%
\begin{abstract}
Forward reachability analysis is a dominant approach for verifying
reach--avoid specifications in neural feedback systems, i.e., dynamical systems
controlled by neural networks, and a number of directions have been proposed and
studied.  In contrast, far less attention has been given to backward
reachability analysis for these systems, in part because of the limited
scalability of known techniques. In this work, we begin to address this gap by
introducing new algorithms for computing both over- and underapproximations of
backward reachable sets for
nonlinear neural feedback systems. We also describe and implement an integration of these backward
reachability techniques with existing ones for forward analysis.  We call the
resulting algorithm
\textbf{F}orward \textbf{a}nd \textbf{B}ackward \textbf{R}eachability \textbf{I}ntegration for \textbf{C}ertification (FaBRIC). We evaluate our
algorithms on a representative set of benchmarks and show that they
significantly outperform the prior state of the art.
\end{abstract}

%%%%%%%%%%%%%%%%%%%%%%%%%%%%%%%%%%%%%%%%%%%%%%%%%%%%%%%%%%%%%%%%%%%%%%%%%%%%%%%
\section{Introduction}
Dynamical systems controlled by neural networks are becoming increasingly
prevalent, with applications in robotics~\cite{csomay2024robust}, autonomous
systems~\cite{ettinger2021large,kaufmann2023champion}, and
more~\cite{wang2021deep}. Such systems are called by various names in the
literature, including
\textit{neural feedback loops}~\cite{rober2023backward} and \textit{neural
  network control systems}~\cite{manzanas_lopez2024arch}.  In this paper, we
follow~\cite{akinwande2025verifying} and refer to such a system 
as a \emph{neural feedback system} (NFS). As more of these systems are
deployed, ensuring compliance with safety and design specifications becomes
increasingly critical. In particular, it is essential to be able to verify that
an NFS satisfies its intended specification under all relevant operating conditions. 

Existing approaches for verifying neural feedback systems include sampling-based methods~\cite{delecki2025diffusion}, certificate synthesis~\cite{mandal2024formally}, and reachability analysis~\cite{akinwande2025verifying}. Sampling, a form of falsification, seeks to identify violations of a specification by evaluating the system over a finite set of scenarios and identifying failure modes~\cite{delecki2025diffusion}. This approach is widely used in industrial practice due to its scalability; however, it cannot be used to provide formal guarantees.

Certificate synthesis methods provide formal guarantees by constructing safety or liveness certificates using techniques such as Lyapunov or Zubov analysis~\cite{9126840,li2025two}, barrier methods~\cite{sha2021synthesizing}, contraction metrics~\cite{sun2021learning}, or combinations thereof~\cite{mandal2024formally}. Systems verified using these approaches have the advantage of being correct by construction, but computing such certificates is often challenging in practice.

Reachability analysis aims to characterize the set of all system states that can be reached to determine whether a given specification is satisfied or violated. While reachability analysis provides formal guarantees and applies to a broad class of systems, it suffers from the curse of dimensionality~\cite{chen2018hamilton}. As a result, state-of-the-art reachability methods seek to balance the trade-off between precision and scalability, with highly precise methods scaling poorly.

Most reachability analysis methods for neural feedback systems focus on forward
analysis, in which the set of possible future states is computed to determine
whether a given NFS satisfies a specification. This focus arises in part
because computing backward reachable sets requires propagating sets backward
through a neural network, a task that is known to be
challenging~\cite{rober2023backward}. The difficulty is further exacerbated
when the NFS has nonlinear transition dynamics, as the analysis must account
for nonlinearity in both the neural network and the system dynamics. 
Even forward analysis of nonlinear neural feedback systems is already quite
challenging~\cite{sidrane2022overt}.

In this work, we explore a verification strategy that combines forward
reachability analysis with new techniques for backward reachability analysis,
with the goal of mitigating some of these scalability limitations. Our contributions are as follows:
\begin{itemize}
	\item We define an algorithm for computing overapproximations of backward reachable sets for nonlinear neural feedback systems using polyhedral enclosures~\cite{akinwande2025verifying}. As part of this algorithm, we extend domain refinement~\cite{everett2023drip} to the nonlinear setting.
	\item We introduce three new algorithms for computing underapproximations of backward reachable sets for nonlinear neural feedback systems. These algorithms explore trade-offs between scalability and precision: one prioritizes scalability, one prioritizes precision, and the third aims to balance both.
	\item We evaluate the proposed backward reachability algorithms on a representative set of benchmarks and demonstrate significant improvements over the prior state of the art for both over- and underapproximation.
	\item We introduce the \textsc{fabric} algorithm, which combines
          forward and backward reachability analysis.  \textsc{fabric} builds
          on existing approaches for combining forward and backward analysis
          but adapts them for the context of neural feedback systems. We
          implement the algorithm and show that it improves the performance of
          reachability analysis for neural feedback systems.
\end{itemize}

The remainder of the paper is organized as follows.  We first present
background material in \Cref{sec:background}. Then, in \Cref{sec:related}, we
review relevant work from the literature.  Next, \Cref{sec:reach analysis}
presents our algorithms for backward reachability analysis, and
\Cref{sec:fabre} describes the \textsc{fabric} algorithm for combining forward
and backward reachability.  We conclude with~\Cref{sec:eval}, which provides a
thorough evaluation of these algorithms.

\section{Background}
\label{sec:background}

\noindent
In this section, we %follow the formalism of \cite{akinwande2025verifying}. 
review notation and definitions used throughout the paper.

\subsection{Notation}
We denote
the set of integers as \ints, the set of natural numbers (integers greater than zero) as \nats, the set of real numbers as \reals, and 
the set of non-negative real numbers as $\reals_+$.
If \stateBig is a set, we denote the \emph{power set} of \stateBig (i.e., the set of all subsets of \stateBig) as $ \powset{\stateBig} $.  We write the Cartesian product of two sets $\stateset$ and $\mathcal{Y}$ as $\stateset \times \mathcal{Y} = \{(x,y) \mid x\in\stateset \wedge y\in\mathcal{Y}\}.$  The $n$-ary Cartesian product is defined analgously, and $\stateset^n$ is used to denote the $n$-ary Cartesian product of the set $\stateset$ with itself.
We use $ [i.. j] $ to represent the set $ \{z\in\ints \mid i\le z \le j\} $ and $[n]$ to abbreviate $[1..n]$.  We write  $[x,y]$ for the set $\{r\in\reals \mid x\le r\le y\}$, $(x,y)$ for the set $\{r\in\reals \mid x < r < y\}$, and define $(x,y]$ and $[x,y)$ in the natural way.

If $\vec{S}$ is any finite \emph{sequence} $ (s_1,\dots,s_n) $, we write $ |\vec{S}| $ to
denote $ n $, the length of the sequence, and $ \vec{S}[i] $ to denote the $i^{\mathit{th}}$ element of the sequence.
We write $\vec{S}[i..j]$ for the sequence $(s_i,\dots,s_j)$  and $ \vec{S} \circ \vec{S}' $ for the sequence obtained by
appending the sequence $ \vec{S}' $ to the end of $ \vec{S} $.
If a sequence is used where a set is expected, the meaning is the set of elements in the sequence (e.g., $s\in \vec{S}$ means that $s$ occurs in the sequence $\vec{S}$).
We use bold face for both vectors and sequences and treat them as interchangeable. 

We use \vleq and \vgeq to describe element-wise inequalities for finite sequences, defined as follows.  If $\vec{x}$ and $\vec{y}$ are sequences of size $n$, then so is $\vec{x} \vleq \vec{y}$, with $(\vec{x} \vleq \vec{y})[i] = 1$, if $\vec{x}[i] \le \vec{y}[i]$, and $0$ otherwise. We define $\vec{x} \vgeq \vec{y}$ similarly. We write $\vec{x} \cdot \vec{y}$ for the dot product
(i.e., sum of element-wise products)
of vectors $\vec{x}$ and $\vec{y}$.  
We write $ \mathbf{0} $ for the zero vector, $ \mathbf{1} $ for the vector of all ones, and $ \unit{i} $ for the $ i^{\mathit{th}} $ unit vector. The sizes of these vectors will be left implicit when it is clear from context. A \emph{convex combination vector} $\vec{\theta}$ is a vector whose entries are non-negative and sum to 1, i.e., $\vec{\theta} \cdot \mathbf{1} = 1$ and $ \vec{\theta} \vgeq \mathbf{0} = \mathbf{1}$.

For a function $ f:\stateBig \to \reals$ and a set $\stateBig' \subseteq \stateBig$, we define the \emph{image} of $\stateBig'$ under $f$ as
$f(\stateBig') := \{f(x) \mid x \in \stateset'\}$.  If $\vec{S}$ is a sequence, then $f(\vec{S})$ is the sequence $(f(\vec{S}[1]),f(\vec{S}[2]),\dots)$.  If $\vec{F}$ is a sequence of functions, then $\vec{F}(x)$ is the sequence $(F[1](x),F[2](x),\dots)$.
For $\stateset' \subseteq \stateset$, we define the \emph{restriction} of $f$ to $\stateset'$ as the function $f^{\stateset'} : \stateset' \to \reals$ such that $f^{\stateset'}(x) = f(x)$ for every $x\in\stateset'$.
%\sa{I don't think we need to discuss switching to a contingency controller tbh}
\subsection{Neural Feedback Systems}
We define a discrete-time neural feedback system $\dynsys$ as the tuple
\[
\langle n, \dom, \init, \trans, \noise, \ctrl, \timestep, \horizon, \reach, \avoid \rangle,
\]
where $n \in \nats$ is the \emph{dimension} of the system (i.e., each state is
an element of $\reals^n$), $\dom \subseteq \reals^n$ is the \emph{operating
domain}---the region over which the system is to certified. We assume that it contains the
the subset of $\reals^n$ in which we expect the system to remain during
operation and that the controller is defined over this region. 
$\init \subseteq \dom$ is the set of \emph{initial states}, 
$\trans = (f_1, \dots, f_n)$ is a sequence of \emph{state transition functions} with $f_i : \reals^n \to \reals$, 
$\noise \subseteq \reals^n$ is the \emph{perturbation set} (i.e., the set of admissible disturbance terms that may affect the next state), 
$\ctrl : \reals^n \to \reals^n$ is the \emph{control function}, 
$\timestep \in \reals_+$ is the \emph{time-step size}, 
$\horizon \in \nats$ is the \emph{number of time steps}, 
$\reach \subseteq \reals^n$ is the set of \emph{goal states}, 
and $\avoid \subseteq \reals^n$ is the set of \emph{avoid states} (i.e., unsafe states).

We assume that states evolve over a sequence of $\horizon$ discrete time steps, each of duration $\timestep$, giving a total time horizon of $\timestep \cdot \horizon$. 
For a current state $\state \in \reals^n$, the next-state function $\nextt^{\dynsys} : \reals^n \to 2^{\reals^n}$ defines the possible successor states of \dynsys. 
This set-valued mapping captures the nondeterminism introduced by the perturbation term:
\begin{align}
    \nextt^{\dynsys}(\state) 
    = \big\{ \state + \big(\vec{F}(\state) + \ctrl(\state) + \vec{\epsilon}\big) \cdot \timestep\ \mid\ \epsilon\in\noise\big\}.
    \label{eq:fwd_diffEq}
\end{align}
Note that when computing the next value of $\state[i]$, $f_i$ and $\ctrl$ may depend on the entire previous state $\state$, not only on the component $\state[i]$. 
We assume that the neural network controller $\ctrl$ is a fully-connected, feed-forward network with $n$ inputs, $n$ outputs, and ReLU activations.
  
The \emph{preimage} (i.e., the set of predecessors) of a set of states $\stateBig$, and must be treated with care due to the presence of nondeterminism. Adversarial perturbations induce two distinct notions of predecessors: those that \emph{must} reach $\stateBig$, and those that \emph{may} reach $\stateBig$. Let $\nextt^{\dynsys}(\cdot, t)$ denote the set of states reachable by $t$ successive applications of the next-state relation. The set of states that \emph{must} reach $\stateBig$ in $t$ steps is defined as
\begin{align}
    \prevg^{\dynsys}(\stateBig, t)
    \;=\;
    \big\{\, \vec{y} \in \reals^n \;\big|\; \nextt^{\dynsys}(\vec{y}, t) \subseteq \stateBig \,\big\}.
    \label{eq:prev_must_t}
\end{align}
That is, every state reachable from $\vec{y}$ in exactly $t$ steps lies in $\stateBig$.
Conversely, the set of states that \emph{may} reach $\stateBig$ in $t$ steps is defined as
\begin{align}
    \preva^{\dynsys}(\stateBig, t)
    \;=\;
    \big\{\, \vec{y} \in \reals^n \;\big|\; \nextt^{\dynsys}(\vec{y}, t) \cap \stateBig \neq \emptyset \,\big\}.
    \label{eq:prev_may_t}
\end{align}
That is, at least one state reachable from $\vec{y}$ in $t$ steps lies in $\stateBig$.
The choice between these previous-state operators depends on the property being verified, and we elaborate on this distinction in the following section.
\subsection{Reach-Avoid Properties}
Let $\stateBig_0$ be a set of initial states.
We define a forward trajectory $\ftraj^{\dynsys}(\stateBig_0)$, as the sequence of state sets $(\stateBig_0, \stateBig_1, \dots, \stateBig_T)$, where $\stateBig_i = \nextt^{\dynsys}(\stateBig_{i-1})$ for all $i \in [1, T]$. 
Similarly, if $\stateBig_T$ is a target, we define the \emph{guaranteed} backward trajectory $\btraj_G^{\dynsys}(\stateBig_T)$ as the sequence of state sets $(\stateBig_T, \stateBig_{T-1}, \dots, \stateBig_0)$, where $\stateBig_i = \prevg^{\dynsys}(\stateBig_T, T-i)$ for all $i \in [0, T-1]$. The \emph{allowable} backward trajectory $\btraj_A^{\dynsys}(\stateBig_T)$ is defined analagously, with $\prevg^\dynsys$ replaced by $\preva^\dynsys$.

We say that a system $\dynsys$ is \emph{safe} if it satisfies either the forward or backward reach-avoid properties. 
The \emph{forward reach-avoid properties} are given by~\eqref{eq:fwd_props}:
\begin{subequations}
\begin{align}
    &\forall\, \state \in \init.\: \exists\, t \in [0..T].\: \ftraj^{\dynsys}(\{\state\})[t] \subseteq \reach, \label{eq:fwd_reach_prop}\\
    &\forall\, t \in [0..T].\: \ftraj^{\dynsys}(\init)[t] \cap \avoid = \emptyset, \label{eq:fwd_avoid_prop}
\end{align}
\label{eq:fwd_props}
\end{subequations}
and the \emph{backward reach-avoid properties} are given by~\eqref{eq:bw_props}:
\begin{subequations}
\begin{align}
    &\exists\, t \in [0..T].\: \btraj_G^{\dynsys}(\reach)[t] \supseteq \init, \label{eq:bw_reach_prop}\\
    &\forall\, t \in [0..T].\: \btraj_A^{\dynsys}(\avoid)[t] \cap \init = \emptyset.
    \label{eq:bw_avoid_prop}
\end{align}
\label{eq:bw_props}
\end{subequations}

\vspace*{-1em}
\noindent
Properties~\eqref{eq:fwd_reach_prop} and~\eqref{eq:bw_reach_prop} are  \emph{reach} properties. 
They ensure that every forward trajectory starting from a state in the initial set eventually reaches the goal set.
Conversely, Properties~\eqref{eq:fwd_avoid_prop} and~\eqref{eq:bw_avoid_prop} are the \emph{avoid} properties, which ensure that all possible system trajectories avoid unsafe states at every time step within the given horizon. 

Properties~\eqref{eq:fwd_props} and~\eqref{eq:bw_props} are each individually \emph{sufficient} to establish the safety of a system, but relying solely on either forward or backward analysis is limited by the challenges discussed earlier.
Our verification strategy jointly leverages~\eqref{eq:fwd_props} and~\eqref{eq:bw_props} to soundly and effectively verify the safety of arbitrary nonlinear neural feedback systems. 

\subsection{Polyhedra}
Let $n,k \in \nats$, let $D \subseteq \reals^n$, and let $P = \{\vec{p}_0, \vec{p}_1,
\dots, \vec{p}_k\} \subseteq D$ be a finite set of points.
The \emph{convex hull} of $P$ is a \emph{polyhedron} defined as:
    \begin{equation}
        \conv(P) 
        = \{\vec{p}_0\, \vec{\theta}[1] + \ldots + \vec{p}_k\, \vec{\theta}[k+1] \mid \vec{\theta} \cdot \vec{1} = 1, \vec{\theta} \vgeq \vec{0} \}.
    \end{equation}
If $\vec{p}_1 - \vec{p}_0,\ \vec{p}_2 - \vec{p}_0,\ \dots,\ \vec{p}_k -
\vec{p}_0$ are linearly independent, $P$ is called \emph{affinely
independent}. If $k = n$, then $P$ is called \emph{full-dimensional}.
A subset of $\reals^n$ is a \emph{polyhedron} if it is the convex hull of some
set of points.
If $\poly$ is a polyhedron, let $ \verts(\poly)$, the \emph{vertices} of $\poly$,
denote the (unique) minimal set of points $\pset$ such that $\poly=\conv(\pset)$.
A $k$-\emph{simplex} is the convex hull of $k+1$ affinely independent points in
$\reals^n$, and $k$ is its \emph{dimension}. If $\simp$ is a $k$-simplex, then
$|\verts(\simp)|=k+1$. A \emph{face} of $\simp$ is the convex hull of any non-empty subset of $\verts(\simp)$.

A \emph{pure simplicial $k$-complex} $\Delta$ is a finite set of simplices such
that: ($i$) the maximum dimension of any simplex in $\Delta$ is $k$; ($ii$)
$S\in\Delta$ iff $S$ is a face of some $S'\in\Delta$; and ($iii$) for
any $\simp_1, \simp_2 \in \Delta$, if $\simp_1 \cap \simp_2 \neq \emptyset$,
then $\simp_1 \cap \simp_2$ is a face of both $\simp_1$ and $\simp_2$.
A pure simplicial complex $\Delta$ is a \emph{triangulation} of $P$ if
\[
\conv(P) \;=\; \bigcup_{\simp \in \Delta} \simp
\quad\text{and}\quad
P \;=\; \bigcup_{\simp \in \Delta} \verts(\simp).
\]

Let $C\subseteq\reals^n$.  The vertices of $C$ with respect to a point set $\pset$ are defined as $\verts_{\pset}(C) = C \cap \pset$.
For a polyhedron $\poly$, the \emph{circumsphere} of $\poly$ (when it exists) is a hypersphere whose surface contains all vertices of $\poly$. Circumspheres always exist for simplices and for hyperrectangles.
Now, let $\pset \subseteq \reals^n$ be a finite, full-dimensional set of
points, let $\Delta$ be a triangulation of $\pset$, and let $\simp$ be an
$n$-simplex in $\Delta$. We define $C(\simp)$ to be the interior of the
$n$-ball whose boundary is the circumsphere of $\simp$. The simplex $\simp$
satisfies the \emph{Delaunay condition}, and is called a \emph{Delaunay simplex
of $\pset$}, if
\[
\verts_{\pset}\!\left(C(\simp)\right) = \emptyset ,
\]
that is, no point of $\pset$ lies in the interior of the circumsphere of $\simp$. The triangulation $\Delta$ is called a \emph{Delaunay triangulation} if every
$n$-simplex in $\Delta$ satisfies the Delaunay condition.

\subsection{Polyhedral Enclosures}
We make use of the polyhedral enclosure abstraction introduced in \cite{akinwande2025verifying}. 
We summarize the relevant definitions and results here and refer readers to that work for details.

Let $n \in \nats$, let $D \subseteq \reals^n$, and let 
$P = \{\vec{p}_0, \vec{p}_1, \dots, \vec{p}_n\} \subseteq D$ be an affinely independent set of points.
Let $L, U : \reals^n \to \reals$ be functions satisfying $L(\vec{p}) \leq U(\vec{p})$ for all $\vec{p} \in P$. 
We call the tuple $B = \langle n, P, L, U \rangle$ a \emph{bounding set}.
The vertex set of $B$ is defined as
\[
    \verts(B) \;\triangleq\; \{\vec{p}\circ(L(\vec{p})) \mid \vec{p} \in P\} \;\cup\; \{\vec{p}\circ(U(\vec{p})) \mid \vec{p} \in P\}.
\]

Let $B$ be a bounding set, and let $\Delta$ be a Delaunay triangulation of $P$, the point set associated with $B$. 
Let $\Delta_n$ denote the set of all  $n$-simplices in $\Delta$. 
For each $S \in \Delta_n$, the \emph{bounding set associated with $S$} is defined as
\[
    B_S \;\triangleq\; \langle n, \mathbf{vert}(S), L^{\mathbf{vert}(S)}, U^{\mathbf{vert}(S)} \rangle.
\] 
The \emph{polyhedron induced by $B_S$} is of dimension $n+1$ and is defined as the convex hull of its vertex set.
\[
    \mathcal{P}(B_S) \;\triangleq\; \mathbf{conv}\!\bigl(\verts(B_S)\bigr).
\]
The \emph{polyhedral enclosure induced by the bounding set $B$ and the Delaunay triangulation $\Delta$} is defined as the union of the bounding polyhedra over all $n$-simplices:
\[
    \mathcal{E}(B,\Delta) \;\triangleq\; \bigcup_{S \in \Delta_n} \mathcal{P}(B_S).
\]
We say that the bounding set $B$ \emph{encloses} a function $f : D \to \reals$
if, for every Delaunay triangulation $\Delta$ and for all $\vec{x} \in \dom$, we have $(\vec{x},f(\vec{x})) \in
\mathcal{E}(B,\Delta)$. Algorithms for constructing such enclosures for a large class of nonlinear functions can be found in~\cite{akinwande2025verifying}.

Given a bounding set $B$ that encloses a function $f$, we can use standard techniques to construct a mixed-integer linear program (MILP) $\mathcal{M}$ with the property that $(\state,y)$ is a solution of $\mathcal{M}$ iff it lies within the polyhedral enclosure induced by $B$ (and some Delaunay triangulation).  Thus, in particular, $(\state,f(\state))$ must be a solution of $\mathcal{M}$.

\section{Related Work}
\label{sec:related}
Reachability analysis computes the set of states that a dynamical system can attain under all admissible control inputs and disturbances \cite{althoff2021set,bansal2017hamilton}. 
In the \emph{forward} reachability problem, one computes the reachable set of states starting from a given initial set over a specified time horizon, whereas the \emph{backward} reachability problem characterizes the set of states from which the system can be driven into a target set. 
Reachability analysis applies to both continuous and hybrid systems
\cite{asarin2006recent}, and it plays a central role in the verification and
control of safety-critical systems%, including applications in air traffic management
~\cite{chen2018hamilton}.

Methods for reachability analysis are commonly classified into \emph{Eulerian} and \emph{Lagrangian} approaches \cite{mitchell2007comparing}. 
Eulerian methods compute reachable sets by discretizing the state space and evolving a value function over a fixed grid \cite{bansal2017hamilton,chen2018hamilton,mitchell2007comparing}, with Hamilton--Jacobi (HJ) reachability serving as a prominent example. 
These methods naturally support backward reachability analysis, and provide strong theoretical guarantees, but they typically incur substantial computational cost due to the need for high-dimensional grid discretizations.

Lagrangian methods compute reachable sets by propagating the system dynamics forward in time from an initial set, typically by enclosing families of trajectories \cite{mitchell2007comparing}. 
Prominent tools implementing Lagrangian reachability include CORA \cite{althoff2016cora,kochdumper2020sparse,kochdumper2023constrained} and Flow* \cite{chen2013flow}. 
Compared to Hamilton--Jacobi methods, Lagrangian approaches generally exhibit improved scalability with respect to the state dimension, but they are not naturally suited to backward reachability analysis.

Computing the exact reachable set is challenging, and state-of-the-art methods often focus on computing sound approximations \cite{althoff2021set}. 
In the forward reachability setting, sound overapproximations are needed, whereas the backward reachability problem typically requires both over and underapproximations. 
The Lagrangian methods discussed above all compute approximations of the reachable set.

\subsubsection{Neural Feedback Systems.}
The verification problem becomes substantially more challenging when the dynamical system is controlled by a neural network, as in neural feedback systems, since the solver must reason about the closed-loop interaction between the continuous dynamics and the neural network controller. 

Methods for verifying neural feedback systems can broadly be classified into propagation methods and combinatorial methods \cite{akinwande2025verifying}. 
Propagation methods verify systems by propagating an abstract domain through the system dynamics using a Lagrangian reachability method, and then propagating the resulting reachable set through the neural network via abstraction propagation \cite{dutta2019reachability,huang2022polar,kochdumper2023open}. 
Combinatorial methods, in contrast, encode both the dynamics and the neural network as combinatorial problems
\cite{sidrane2022overt,siefert2023successor,vincent2021reachable}. 
Propagation methods typically scale to higher-dimensional systems at the cost of precision, whereas combinatorial methods retain higher precision but incur substantially greater computational cost. 

Recent work has also explored computing backward reachable sets for neural
feedback systems, with the majority of approaches focusing on systems with
linear transition dynamics
\cite{rober2022backward,kotha2023provably}. Approaches that consider systems
with nonlinear transition dynamics tend to scale poorly
\cite{rober2023backward}. Everett et al.~\cite{everett2023drip} minimize some
of the limitations of backward reachability analysis of neural feedback systems
by introducing a domain refinement scheme. 

\subsubsection{Neural Network Verification.}
Neural network verification aims to characterize the input--output behavior of a neural network in order to establish properties such as robustness and stability. 
Common verification approaches include formulating the problem as a satisfiability modulo theories (SMT) query \cite{katz2017reluplex,katz2019marabou}, a MILP \cite{tjeng2017evaluating}, or an abstraction propagation problem \cite{singh2019abstract,zhang2018efficient,wang2021beta}. 
Computing the pre-image of a neural network is known to be particularly challenging \cite{zhang2025premap}, and recent work has therefore focused on constructing sound overapproximations \cite{kotha2023provably} and underapproximations \cite{zhang2024provable} of neural network pre-images.

Forward and backward analysis have been explored as mechanisms for bound tightening in neural network verification. 
The authors of \cite{wang2021beta} implicitly combine forward and backward analysis to tighten neuron bounds during interval propagation, while the authors of \cite{wu2022scalable} explicitly combine forward and backward analysis to verify graph neural networks. 
Forward and backward analysis have also been studied in the context of reachability analysis \cite{mitchell2007comparing}, but scalable computation of backward reachable sets remains a fundamental challenge. 
Recent work explores the use of forward and backward reachability analysis for domain refinement in \emph{linear} neural feedback systems \cite{wang2025verifying}. 
\section{Backward Reachability Analysis}
\label{sec:reach analysis}
In prior work~\cite{rober2023backward}, \emph{backward reachable sets} are defined as the sets of states that reach a target under \emph{any} admissible control, while \emph{backward projection sets} are those that reach the target under the specific control policy implemented by the neural network. This distinction is essential in~\cite{rober2023backward} because the proposed methods require explicit knowledge of bounds on the admissible control inputs. In contrast, our setting does not impose such a requirement. Accordingly, we use \emph{backward reachable set} to denote the set of states that reach a target set under the specific neural network controller policy.  We further refine this terminology based on two (\emph{must} vs \emph{may}) notions of backward reachable sets.  Specifically, we call a solution to Equation~\eqref{eq:prev_may_t} a \emph{backward may-reachable set} or \emph{may-BRS} and a solution to Equation~\eqref{eq:prev_must_t} a \emph{backward must-reachable set} or \emph{must-BRS}.

We consider nonlinear neural feedback systems, and therefore rely on approximate solutions to \Cref{eq:fwd_diffEq,eq:prev_must_t,eq:prev_may_t}. For forward reachability analysis, an overapproximate solution to \Cref{eq:fwd_diffEq} does not compromise the soundness of Properties~\eqref{eq:fwd_reach_prop} and \eqref{eq:fwd_avoid_prop}. For backward analysis, however, an overapproximate solution to~\Cref{eq:prev_may_t} can only be used with Property~\eqref{eq:bw_avoid_prop}. To maintain the soundness of Property \eqref{eq:bw_reach_prop}, an underapproximate solution to~\Cref{eq:prev_must_t} is required. For ease of exposition, we refer to overapproximations of backward may-reachable sets as \emph{outer sets} or \emph{outer approximations} and underapproximations of backward must-reachable sets as \emph{inner sets} or \emph{inner approximations}. In the following subsections, we describe methods for computing both outer and inner approximations.

\subsection{Outer Sets}
Our objective is to develop algorithms that soundly compute overapproximations of the solutions to~\eqref{eq:prev_may_t} in a tractable manner.
Inspired by the procedure described in~\cite{rober2023backward}, we formulate
the following problem. Given a NFS~$\dynsys$ with operating domain~\dom and target
set~\stateset, we compute bounds on the solution
of~\eqref{eq:prev_may_t} by solving a series of mixed-integer linear programs.
We first employ the technique of~\cite{akinwande2025verifying} to obtain sound polyhedral enclosures for the nonlinear transition dynamics.
Then, for each $i \in [n]$, we solve the following MILP in order to get a lower
bound on $\state[i]$.
\begin{subequations}
\begin{align}
    \min \; \state[i], \\
    \text{subject to } 
    \quad \state' = \state + \big(\vec{y} + \ctrl + \vec{\epsilon}\big)\cdot\timestep, \\
    \quad \model_0,\dots,\model_n, \quad
    \vec{\epsilon} \in \noise, \quad
    \state' \in \stateset, \quad
    \state \in \dom
\end{align}
\label{eq:bw_Reach_outer}
\end{subequations}

\vspace*{-1em}
\noindent
where $\model_1,\dots,\model_n$ are MILPs encoding the polyhedral enclosures for $f_1,\dots,f_n$, respectively, and whose input and output variables are $(\state,\vec{y}[1]), \dots, (\state,\vec{y}[n])$, and $\model_0$ is a MILP formulation of the fully-connected ReLU-activated neural network controller using standard methods~\cite{tjeng2017evaluating}, whose input and output variables are $(\state,\vec{u})$. Replacing the minimization with a maximization yields corresponding upper bounds. Together, these bounds define a hyperrectangle that is an outer set, a sound overapproximation of the true may-BRS of \stateset.

This outer set is sound but conservative. The conservatism arises from the fact that we typically have extremely loose bounds on $\state$: for soundness, we use the domain $\dom$ to bound $\state$, but $\dom$ may substantially overapproximate the set of states that are actually backward reachable. We address this next.

\subsubsection{Extending Domain Refinement to Nonlinear Neural Feedback Systems.}
The original method introduced in~\cite{rober2023backward} mitigates this issue by assuming \emph{a priori} bounds on the controller outputs and using these bounds to compute a restricted subset of the domain over which the backward reachability problem is solved. However, this relies on obtaining tight controller bounds, which is often difficult in practice. When such bounds are loose or unavailable, the resulting overapproximate backward reachable set remains highly conservative.

Subsequent work~\cite{everett2023drip} eliminates the need for \emph{a priori} controller bounds by proposing a domain refinement algorithm that iteratively tightens the domain over which the neural feedback system is defined. That algorithm is formulated for neural feedback systems with linear transition functions. A minor contribution of this work is to extend their domain refinement scheme to systems with nonlinear transition functions. %We describe the modified refinement procedure below.

The refinement algorithm introduced in~\cite{everett2023drip}, referred to as \emph{DRIP}, forms the basis of our approach. The central insight underlying DRIP is that increasingly tight overapproximations of the backward reachable set can be obtained by starting with an overly conservative approximation and iteratively refining it---using the previously computed overapproximate backward reachable set as the domain constraint at each iteration. DRIP uses the CROWN framework~\cite{zhang2018efficient} to compute the backwards reachable set. While effective for systems with linear transition functions, i.e., systems whose dynamics can be encoded as linear programs, this approach cannot be directly applied to nonlinear transition systems. To address this limitation, we extend DRIP to systems with nonlinear transition functions as shown in \Cref{alg:drip}.% We retain the DRIP acronym to emphasize continuity with the original method. The resulting algorithm is presented below.

\begin{algorithm}
\caption{Domain Refinement with Polyhedral Enclosures (DRiPy)}
\label{alg:drip}
\begin{algorithmic}[1]
\Require Target set $\stateset$, 
         neural feedback system $\dynsys$,
         maximum number of iterations $K_{\max}$,
         improvement threshold $\alpha$.
\Ensure overapproximation of the one-step may-BRS $\preva^{\dynsys}(\stateset, 1)$
\State $k \gets 0$ \Comment{Initialize iteration counter}
\Repeat
    \State $\mathcal{M}_0 \gets \textsc{ApproxNetwork}(\vec{\ctrl},\dom)$  \Comment{Construct MILP approximating controller $\vec{u}$}
    \State $\mathcal{M}_1,\dots,\mathcal{M}_n \gets
    \textsc{OvertPoly}(\vec{\trans}, \dom)$ \Comment{Compute MILPs for nonlinear dynamics}
    \State $(\vec{L}, \vec{U}) \gets
    \textsc{Solve}(\dynsys,\mathcal{M}_0,\dots,\mathcal{M}_n)$ \Comment{Solve
      \eqref{eq:bw_Reach_outer} to get bounds on may-BRS}
    \State $\dom_{\mathit{old}} \gets \dom$ \Comment{Save domain $\dom$}
    \State $\dom \gets \textsc{Hyperrectangle}(\vec{L}, \vec{U})$ \Comment{Refine $\dom$ to the
      hyperrectangle formed by $\vec{L}$ and $\vec{U}$}
    \State $k \gets k + 1$ \Comment{Increment iteration counter}
\Until{ $\mathit{vol}(\dom_{\mathit{old}}) / \mathit{vol}(\dom) - 1 < \alpha$ \textbf{ or } $k \ge K_{\max}$ }
\State \Return $D$
\end{algorithmic}
\end{algorithm}

\vspace*{-1em}
\noindent

\Cref{alg:drip} begins by computing the MILPs that encode an overapproximation
of $\ctrl$ and the polyhedral enclosures of each $f_i\in\trans$ using
techniques from~\cite{akinwande2025verifying}.  It then solves $n$ instances of
\eqref{eq:bw_Reach_outer} to get a vector of lower bounds $\vec{L}$ and solves
$n$ instances of the maximization (replacing $\min$ by $\max$) to get a vector of
upper bounds $\vec{U}$.  These are used to create a new hyperrectangle
which is assigned to be the new domain $\dom$.  The process then repeats until
the improvement in the volume falls below some threshold $\alpha$ or the maximum
number of iterations $K_{\max}$ has been reached.

\Cref{alg:drip} can be extended in the obvious way to $k$ time steps by simply calling the full algorithm $k$ times, each time replacing the target set $\stateset$ by the overapproximated may-BRS $D$ calculated in the previous call.
Borrowing terminology from the forward reachability literature~\cite{sidrane2022overt}, we refer to this as \emph{concrete} overapproximate backward reachability analysis.

The concrete approach can induce excessive conservatism, mirroring behavior observed in concrete forward reachability analysis \cite{akinwande2025verifying}. To mitigate this effect, we extend the symbolic multi-step approach of~\cite{akinwande2025verifying,sidrane2022overt} by solving for multiple time steps simultaneously.
We define the two-step problem as follows.  We solve, for each $i \in [n]$:
\begin{subequations}
\begin{align}
 \min \; \state[i],\\
 \text{subject to }
 \begin{cases}
   \state'' = \state' + \big(\vec{y'} + \ctrl' + \vec{\epsilon'} \big)\cdot\timestep \\
   \state' = \state + \big(\vec{y} + \ctrl + \vec{\epsilon}\big)\cdot\timestep
 \end{cases}&\\
 \model_0,\dots,\model_n, \quad
  \model'_0,\dots,\model'_n, \quad
  \vec{\epsilon},\vec{\epsilon'}\in\noise\quad
  \state'' \in \stateset, \quad
  \state', \state \in \dom
\end{align}
\label{eq:sym_bw_Reach_Outer}
\end{subequations}

\vspace*{-1em}
\noindent
where $\model'_1,\dots,\model'_n$ are the MILPs encoding the polyhedral enclosures for $f_1,\dots,f_n$ at the transition before the target is reached, $\model_1,\dots\model_n$ are the MILPs encoding the polyhedral enclosures at the transition before that one, and the terms $\model'_0$ and $\model_0$ denote the MILP formulations of the neural network controller at the previous transition and the one before it, respectively.
As before, we can compute an upper bound by replacing the minimization with a maximization. We can also run \Cref{alg:drip} with the new formulation in \eqref{eq:sym_bw_Reach_Outer} to refine the two-step backward may-reachable set. Finally, we can implement a \emph{hybrid} strategy by first running the concrete approach, then replacing the domain $D$ with the computed concrete BRS at each stage to improve precision.

\subsection{Inner Sets}

%Intuition: starting with an outer set is helpful becuase it gives you a smaller search space to deal with, and we know the inner set has to be in there

By definition, any sound outer set must contain the true backward reachable set. We exploit this property to compute an underapproximation of the true must-BRS by formulating a constrained optimization problem whose objective is to identify the largest subset of the outer set whose overapproximated forward image is fully contained in the target set. This leads to the following optimization problem:
\begin{subequations}
\begin{align}
    &\max \;\textsc{vol}(\stateset^i), \\
    &\text{subject to} \quad 
      \textsc{Reach}(\stateBig^i) \subseteq \stateset, \\
    & \qquad\qquad\quad 
      \stateBig^i \subseteq \stateBig^o,
\end{align}
\label{eq:bw_Reach_inner}
\end{subequations}

\vspace*{-1em}
\noindent 
where $\textsc{vol}$ denotes the volume operator, $\textsc{Reach}$ denotes a sound algorithm for overapproximate forward reachability analysis, $\stateBig^o$ is an outer set, and $\stateset$ is the target set. We next propose several approaches for solving this efficiently. All proposed approaches assume that the user can compute an outer set, as well as an efficient overapproximation of the forward reachable set.

\subsubsection{Scaled Hyperrectangular Approximation of Reachable Polytopes (\ssharp)}
The \ssharp approach computes an inner set by scaling an outer set inward. We assume that the outer set can be represented as $\stateBig^o = \vec{c} \pm \vec{r}$. The inner set is defined as $\stateBig^i = \vec{c} \pm \vec{\rho} \cdot \vec{r}$, where $\vec{\rho} \in (0,1)^n$. Our objective is to estimate the largest admissible scaling vector $\vec{\rho}$ such that $\textsc{Reach}(\stateBig^i)$ is fully contained in the target set. This leads to the following optimization problem:
\begin{subequations}
\begin{align}
    &\max \; \sum_{j=1}^n \log(\vec{\rho}[j]) \\
    &\text{subject to} \quad 
      \mathbf{0} \vleq \vec{\rho} \vleq \mathbf{1}, \\
%    & \qquad\qquad\quad
%      \stateBig^o  = \vec{c} \pm \vec{r}, \\
    & \qquad\qquad\quad 
      \stateBig^i  = \vec{c} \pm \vec{\rho} \cdot \vec{r}, \\
    & \qquad\qquad\quad
      \textsc{Reach}(\stateBig^i) \subseteq \mathcal{\stateset}.
\end{align}
\label{eq:bw_reach_sharp}
\end{subequations}

\vspace*{-1em}
\noindent
We estimate the optimal $\vec{\rho}$ via an iterative line-search procedure inspired by golden-section search. We use golden-section search because it typically requires fewer containment queries than a binary search for comparable bracketing accuracy. Solving equation \eqref{eq:bw_reach_sharp} returns a \emph{certified} inner approximation of the true must-BRS. 

This formulation has several limitations, the most significant of which are the need for frequent containment queries and the assumption that the inner set is an axis-aligned hyperrectangle centered at the center of the outer set. Nonetheless, the approach often converges quickly in practice. The formulation also typically requires a large number of forward reachability queries, which can be problematic when such queries are computationally expensive. To mitigate these limitations, we propose the following alternatives.
\subsubsection{Convex Rectangle Inferred via Sampled Positives (\crisp)}
The \crisp approach seeks to reduce the number of forward reachability queries required to compute an inner set. To this end, we replace the scaling-based procedure with a sampling-based alternative. 
We draw dense, space-filling samples from the outer set. Each sample is then classified as positive or negative via rejection sampling, based on containment of its forward reachable set in the target set.
Let $P$ be the set of positive samples. Our objective is to identify the largest axis-aligned hyperrectangle contained within the convex hull of the positive samples.  The resulting optimization problem is:
\begin{subequations}
\begin{align}
\max \quad & \textsc{vol}(\stateset^i) \\
\text{subject to}\quad
&\stateset^i(\vec{c},\vec{r}) \subseteq \conv(P), \\
&\vec{r} \vgeq \vec{0}.
\end{align}
\label{eq:bw_reach_crisp}
\end{subequations}

\vspace*{-1em}
\noindent
where $\stateset^i(\vec{c},\vec{r})$ is an axis-aligned hyperrectangle. Note that this hyperrectangle may not necessarily share a center with the outer set. With an appropriate surrogate for the $\textsc{vol}$ operator, this optimization problem can be formulated as a convex program. 

While this formulation addresses several of the major limitations of the \ssharp approach, it remains sensitive to irregular backward reachable sets. In particular, when the backward reachable set is nonconvex or contains holes (e.g., a toroidal set), the resulting candidates may fail the forward reachability certification step and therefore be nonviable. We mitigate this limitation via the following modification.
\subsubsection{Constrained Local Exclusion of Aligned Negatives (\clean)}
The \clean approach is designed to handle problem instances with irregular backward reachable sets. Although such cases are uncommon in practice, we include this method for completeness. We employ the same space-filling rejection sampling algorithm used in \crisp, but here we retain the set of negative samples $N$. Our objective is to identify the largest axis-aligned hyperrectangle that contains no negative samples.
\begin{subequations}
\begin{align}
\max \quad & \textsc{vol}(\stateset^i) \\
\text{subject to}\quad
& \stateset^i(\vec{c},\vec{r}) \cap N = \emptyset, \label{subeq:empty}\\
& \stateset^i(\vec{c},\vec{r}) \subseteq \stateBig^o, \\
& \vec{r} \vgeq \vec{0}.
\end{align}
\label{eq:bw_reach_clean}
\end{subequations}

\vspace*{-1em}
\noindent
The computational complexity of this problem depends on both the surrogate used for the volume objective and the representation of the emptiness constraint in~\cref{subeq:empty}. While a convex surrogate may be used for the volume objective, enforcing the emptiness constraint requires integer variables. In particular, binary variables are needed to enforce exclusion along at least one dimension for each negative sample $n \in N$. Consequently, the resulting optimization problem is, in the general case, a mixed-integer linear program. The number of binary variables scales with both the system dimension and the number of negative samples.

Solving the optimization problem in equation \eqref{eq:bw_reach_crisp} or \eqref{eq:bw_reach_clean} yields a \emph{candidate} underapproximate solution to the must-BRS. These optimization problems are used as subroutines in \Cref{alg:fits} for computing an inner set.
\begin{algorithm}
\caption{Fast Inner Template Sets (FITS)}
\label{alg:fits}
\begin{algorithmic}[1]
\Require Target set $\stateset$, neural feedback system $\dynsys$, outer set $\stateset^o$, maximum number of iterations $K_{\max}$, inner approach $p \in \{\textsc{clean},\textsc{crisp}\}$
\Ensure underapproximation of the true must-BRS $\prevg(\stateset,1)$
\State $k \gets 0$ \Comment{Initialize iteration counter}
\Repeat
    % \State $\model_1,\dots,\model_n \gets \textsc{OvertPoly}(\trans,D)$ \Comment{Construct MILPs for nonlinear dynamics}
    \State $\textsc{Inner} \gets \textsc{Select}(p)$ \Comment{Select inner approximation routine}
    \State $\stateset^i \gets \textsc{Inner}(\dynsys,\stateset^o)$ \Comment{Approximate the solution of \eqref{eq:bw_Reach_inner}}
    \If{$\textsc{reach}(\stateset^i) \subseteq \stateset$}
        \State \Return $\stateset^i$ \Comment{Accept candidate if forward query passes}
    \EndIf
    \State $k \gets k + 1$ \Comment{Increment iteration counter}
    \State $\stateset^o \gets \stateset^i$ \Comment{Update outer set}
\Until{$k \ge K_{\max}$}
\State \Return $\emptyset$
\end{algorithmic}
\end{algorithm}
\noindent
We can further optimize \Cref{alg:fits} by sampling a finite set of points from the candidate set $\stateBig^{i}$ and iteratively reducing the set until no negative samples remain \emph{before} performing the forward reachability query.

As in the overapproximation case, the algorithm can be extended to $k$ time steps by
calling the algorithm $k$ times, each time replacing the target set $\stateset$ by the underapproximated must-BRS $\stateset^i$ calculated in the previous call.
We refer to the resulting procedure as a \emph{concrete} underapproximation if candidate sets are validated using a concrete forward reachability query, and as \emph{symbolic} underapproximation if validation is instead performed using a symbolic query.

\section{The \fab Verification Strategy}
\label{sec:fabre}
Rather than exclusively computing forward or backward reachable sets, we propose verifying $\dynsys$ by combining both analyses. 
Specifically, we partition the planning horizon into $\horizon = F + B$ steps, where $F$ denotes the number of forward-analysis steps and $B$ the number of backward-analysis steps. This yields the \textbf{F}orward \textbf{a}nd \textbf{B}ackward \textbf{R}eachability \textbf{I}ntegration for \textbf{C}ertification \textsc{FaBRIC} strategy.

The subdivision is treated as a configurable parameter of the strategy. We generally expect $F > B$ because the forward problem is implicitly encoded within the backward problem; consequently, the computational effort required to perform $k \in \nats$ steps of backward analysis typically exceeds that required for $k$ steps of forward analysis. This discrepancy is even more pronounced when verifying reach properties, since the repeated certification queries introduce additional overhead. In this work, we consider subdivisions satisfying $F \ge 0.75\,\horizon$ for all problems and leave the meta-optimization of $F$ and $B$ to future work.

To establish that the system satisfies a reach property (Properties~\ref{eq:fwd_reach_prop} and~\ref{eq:bw_reach_prop}), we perform $F$ steps of forward analysis from $\init$ and $B$ steps of backward analysis from $\reach$. 
If the overapproximation of the forward reachable set at time step $F$ is contained within the underapproximation of the backward must-reachable set, then the system is guaranteed to reach the goal set. 

To verify that the system satisfies an avoid property (Properties~\ref{eq:fwd_avoid_prop} and~\ref{eq:bw_avoid_prop}), we compute $F$ steps of overapproximated forward analysis from the initial set and $B$ steps of overapproximated backward analysis from the avoid set $\avoid$. Let $U = \cup_{t = 0}^B \preva(A)$.
If each of the forward sets at time $t\in[0,F]$ is disjoint from $A$ and the forward set at time $F$ is additionally disjoint from $U$, then the system is guaranteed to avoid all unsafe states, thereby ensuring safety.
\section{Evaluation}
\label{sec:eval}
We evaluate our algorithms and overall strategy on three classes of neural feedback systems drawn from the ARCH Competition benchmark suite \cite{manzanas_lopez2024arch}. For each benchmark, we consider controllers inspired by those used in the original ARCH Competition, which are obtained by training neural policies via imitation learning from a model predictive control (MPC) expert. 

For each system, we train policies with $0.5\times$, $1\times$, and $2\times$ the number of neurons used in the corresponding ARCH benchmark controller, while retaining all other architectural features. We refer to these variants as the small, medium, and large policies respectively. We provide summaries below, and detailed descriptions in the appendix.
\subsection{Benchmarks}
\paragraph{TORA}
The objective is to verify that the states of an actuated cart remain within a safe region during a specified time interval. 
\paragraph{Unicycle Car Model}
The objective is to verify that a simplified model of an autonomous vehicle is safe. The car is modeled with four state variables, representing the $ x $ and $ y $ coordinates in a plane, the steering angle, and the velocity magnitude.
\paragraph{Attitude Control Model}
The objective is to verify that a simplified model of an autonomous aircraft is safe. The aircraft is modeled as a rigid body with six state variables, representing the orientation ($\psi_1,\psi_2,\psi_3$) using Rodrigues parameters, and rotation rates ($\omega_1,\omega_2,\omega_3$) of the aircraft.

\subsection{Baselines}
We compare our method for computing outer approximations against HyBreach-MILP, a state-of-the-art approach for computing outer reachable sets of nonlinear neural feedback systems~\cite{rober2023backward}. For HyBreach-MILP, we compute bounds on the admissible control inputs via parameter search and report results obtained using the two-step procedure described in~\cite{rober2023backward}.
We compare our method for computing inner approximations against BURNS, the only existing tool for computing inner reachable sets for nonlinear neural feedback systems~\cite{sidrane2025burns}. We reimplement the algorithm as described in the paper and report the resulting comparisons below.
Finally, we compare the \fab strategy against the forward reachability analysis method introduced in~\cite{akinwande2025verifying}\footnote{We use an existing public implementation provided by the authors of~\cite{akinwande2025verifying}.}
. This algorithm represents the state of the art in combinatorial forward reachability for nonlinear neural feedback systems and therefore provides a strong baseline for evaluating our approach.

\subsection{Implementation Details}
The convexity of equations~\eqref{eq:bw_reach_crisp} and \eqref{eq:bw_reach_clean} depends on the surrogates used to compute the \textsc{vol} operator. For simplicity, we assume that the reachable sets are axis-aligned hyperrectangles and define the volume surrogate as the sum of the radii $\vec{r}$. An alternative is to compute the volume using the log-sum (equivalently, the product) of the radii; however, this choice leads to a cone program, which is generally more computationally expensive.
We generate samples using deterministic Sobol sequences~\cite{sobol1967distribution}. This provides an efficient space-filling sampling scheme suitable for our rejection-sampling procedure. We use the Gurobi solver \cite{gurobi} for all optimization queries.

\subsection{Experimental Setup}
All computation times were obtained by running each tool on an AMD Ryzen~9 7950X processor. For soundness, all reported results correspond to certified bounds on the optimal value of the objective function. Since all methods rely on the \textsc{OVERT} algorithm, we use a fixed parameter configuration with $N = 2$, a relative error tolerance of $5 \times 10^{-3}$, and an absolute error tolerance of $10^{-4}$ across all experiments. For the Attitude benchmark, we instead use $N = 1$ for both HyBreach-MILP and BURNS, as these tools otherwise time out.

For all backward reachability benchmarks, we perform five steps of symbolic backward reachability analysis using the corresponding avoid and goal sets. For the overapproximation benchmarks, we impose a per–solver-call timeout of $900$ seconds for each tool, set $K_{\text{max}} = 10$, and use $\alpha = 0.01$. 
For the underapproximation benchmarks, we set $K_{\text{max}} = 10$ and report the best result across $10$ independent attempts for both \ssharp and BURNS. We impose a per–solver-call timeout of $600$ seconds for each algorithm, together with an additional timeout of $300$ seconds for forward reachability queries.

For the \fab benchmarks, we employ a combination of concrete and symbolic reachability for the forward-analysis component, together with purely symbolic reachability for the backward component. 
For the TORA benchmark with $\horizon^{\tora} = 20$, we set $F = 15$ and $B = 5$, and further subdivide the forward component into two hybrid symbolic runs with horizons $T = 10$ and $T = 5$, respectively. 
For the Unicycle benchmark with $\horizon^{\uni} = 50$, we set $F = 47$ and $B = 3$, further subdividing the forward component into four hybrid symbolic runs with horizon $T = 10$ and one run with horizon $T = 7$. 
For the Attitude benchmark with $\horizon^{\att} = 10$, we set $F = 8$ and $B = 2$. We impose a per-solver call timeout of $600$ seconds for all forward and backward analyses.

\section{Results}
\subsection{Computing Outer Sets}
We report results comparing HyBReach-MILP with the \dripy algorithm. To enable a fair comparison, we present results for \dripy both with and without domain refinement. We additionally report the improvement in computation time and set volume achieved by \dripy relative to HyBReach-MILP when no refinement steps are applied.

The TORA benchmark is the easiest among the considered benchmarks. All tools compute outer approximations in under $10$ seconds, with HyBReach-MILP outperforming \dripy in terms of computation time. HyBReach-MILP also produces a tighter outer approximation than unrefined \dripy, which is consistent with expectations: HyBReach-MILP employs a more precise, albeit less scalable, representation of the nonlinear transition functions. Despite this advantage, \dripy with domain refinement computes an outer set that is $3.5\times$ smaller in under $10$ seconds.

The Unicycle and Attitude benchmarks exhibit a different performance trend. In these cases, the computational cost of HyBReach-MILP outweighs its precision advantage, and unrefined \dripy becomes the fastest method. In some instances, unrefined \dripy is nearly two orders of magnitude faster than HyBReach-MILP, while still achieving an order-of-magnitude reduction in outer set volume. These results are partly explained by the timeout imposed on all tools: HyBReach-MILP frequently times out, resulting in larger cumulative computation times and looser outer sets, since we report objective bounds. The medium Unicycle benchmark further highlights the benefits of domain refinement. In this setting, \dripy requires substantially fewer refinement iterations to converge, resulting in significantly lower computation times compared to other variants.

For the Unicycle benchmark, domain refinement reduces the outer-set volume by a factor of $3.5$ relative to unrefined \dripy. Compared to HyBReach-MILP, domain refinement produces an outer set that is over $40\times$ smaller while requiring less computation time. 
For the Attitude benchmark, domain refinement yields an additional reduction factor of approximately $200$ relative to unrefined \dripy. Compared to HyBReach-MILP, domain refinement produces an outer set that is over $2900\times$ smaller while also requiring less computation time.
\begin{table*}[h!]
    \centering
    \resizebox{\columnwidth}{!}{%
    \begin{tabular}{lcccccc|cc}
        \toprule
        & \multicolumn{2}{c}{HyBReach-MILP}
        & \multicolumn{2}{c}{\dripy (\ding{72})}
        & \multicolumn{2}{c}{\dripy (\ding{72}$\ddagger$)}
        & \multicolumn{2}{c}{Improvement} \\
        \cmidrule(lr){2-3}
        \cmidrule(lr){4-5}
        \cmidrule(lr){6-7}
        \cmidrule(lr){8-9}
        & Time (s) & Vol. 
        & Time (s) & Vol.
        & Time (s) & Vol.
        & Time ($\downarrow$) & Vol. ($\downarrow$) \\
        \midrule
        Tora (sma.)  & $\mathbf{1.14}$ & $1.44\textsc{E-01}$ & $1.63$ & $1.67\textsc{E-01}$ & $5.17$ & $\mathbf{4.16\textsc{E-02}}$ & $-1.43$ & $-1.16$ \\
        Tora (med.)  & $\mathbf{1.14}$ & $1.39\textsc{E-01}$ & $2.63$ & $1.57\textsc{E-01}$ & $5.20$ & $\mathbf{4.11\textsc{E-02}}$ & $-2.22$ & $-1.13$ \\
        Tora (larg.)  & $\mathbf{1.40}$ & $1.44\textsc{E-01}$ & $2.53$ & $1.66\textsc{E-01}$ & $6.68$ & $\mathbf{4.16\textsc{E-02}}$ & $-1.81$ & $-1.15$ \\
        \midrule
        Unicycle (sma.)  & $5501.0$ & $1.84\textsc{E+03}$ & $\mathbf{60.59}$ & $1.51\textsc{E+02}$ & $1897.94$ & $\mathbf{4.00}\textsc{E+01}$ & $90.79$ & $12.23$ \\
        Unicycle (med.)  & $4980.90$  & $1.70\textsc{E+03}$ & $\mathbf{55.04}$ & $1.51\textsc{E+02}$ & $124.913$ & $\mathbf{4.20}\textsc{E+01}$ & $90.50$ & $11.30$ \\
        Unicycle (larg.) & $4739.67$ & $1.82\textsc{E+03}$ & $\mathbf{100.58}$ & $1.51\textsc{E+02}$ & $1611.40$ & $\mathbf{4.20}\textsc{E+01}$ & $47.12$ & $12.06$ \\
        \midrule
        Attitude (sma.)  & $1871.68$ & $3.73\textsc{E+05}$ & $\mathbf{77.75}$ & $2.57\textsc{E+04}$ & $1165.71$ & $\mathbf{1.25}\textsc{E+02}$ & $24.07$ & $14.51$ \\
        Attitude (med.)  & $1902.81$  & $3.73\textsc{E+05}$ & $\mathbf{91.46}$ & $2.57\textsc{E+04}$ & $1060.17$ & $\mathbf{1.25}\textsc{E+02}$ & $20.81$ & $14.51$ \\
        Attitude (larg.) & $2256.76$ & $3.73\textsc{E+05}$ & $\mathbf{145.53}$ & $2.57\textsc{E+04}$ & $626.70$ & $\mathbf{1.25}\textsc{E+02}$ & $15.51$ & $14.52$ \\
        \bottomrule
    \end{tabular}}
    \setlength{\abovecaptionskip}{10pt}
     \caption{Benchmark computation time (s) and set volumes. \ding{72} denotes our polyhedral-enclosure based algorithm to solve Eq. \ref{eq:sym_bw_Reach_Outer}, and $\ddagger$ denotes our algorithm combined with the modified refinement scheme. Improvement is calculated as the time and volume fraction between HyBReach-MILP and a configuration of \dripy without domain refinement.}
    \label{tab:results-transposed1}
\end{table*}
\vspace*{-4em}
\subsection{Computing Inner Sets}
We report results comparing BURNS with  \ssharp and variants of the FITS algorithm. By construction, the sets returned by our methods (\ssharp, \crisp, \clean) are certified inner sets, whereas the sets returned by BURNS are not certified. Instead, the sets returned by BURNS are assumed to be certified by construction.

On the TORA benchmark, BURNS outperforms all variants of our algorithm in terms of inner set volume, although we are able to return certified inner sets in a fraction of the computation time. The sets returned by BURNS are nearly $2\times$ larger than the largest set returned by \ssharp or FITS.

This trend does not persist on the more challenging benchmarks. On these benchmarks, BURNS is unable to compute a valid inner set within $K_{\text{max}}$ attempts, whereas all variants of the \textsc{FITS} algorithm succeed. 
For the Unicycle benchmark, BURNS completes execution but is unable to return an inner set with nonzero volume. In contrast, for the Attitude benchmark, BURNS struggles to identify feasible points in the inner set. This difficulty arises largely from the need to sample a sparse set of points in a six-dimensional domain. Our reported results use up to $5 \times 10^{7}$ samples; we were unable to use larger sample counts due to memory limitations.
Our sampling-based approaches avoid this limitation by sampling from the outer set, which constitutes a substantially smaller subset of the domain. Notably, the \crisp algorithm is the best-performing \textsc{FITS} variant across all benchmarks, while the \clean algorithm produces the largest inner set on a single instance. The \ssharp algorithm struggles across all benchmarks but consistently returns a valid, albeit small, inner set.

\begin{table*}[h!]
    \centering
    \resizebox{\columnwidth}{!}{%
    \begin{tabular}{lcccccccc}
        \toprule
        & \multicolumn{2}{c}{BURNS} & \multicolumn{2}{c}{\ssharp(\ding{72})} & \multicolumn{2}{c}{\crisp (\ding{72}) } &\multicolumn{2}{c}{\clean (\ding{72}) }\\ 
        \cmidrule(lr){2-3} \cmidrule(lr){4-5} \cmidrule(lr){6-7} \cmidrule(lr){8-9}
        & Time (s) & Vol. & Time (s)  & Vol. & Time (s) & Vol. & Time (s) & Vol.\\ 
        Tora (sma.) & $32.59$ & $\mathbf{6.35\textsc{E-04}}$ & $\mathbf{6.62}$&$1.85$\textsc{E-05} & $8.52$ & $3.82\textsc{E-04}$ & $6.82$ & $2.41\textsc{E-06}$ \\
        Tora (med.) & $32.59$ & $\mathbf{6.35}\textsc{E-04}$ & $\mathbf{6.72}$&$1.85$\textsc{E-05} & $8.83$ & $3.82\textsc{E-04}$ & $6.79$ & $2.40\textsc{E-06}$ \\
        Tora (lar.) & $65.92$ & $\mathbf{4.70}\textsc{E-04}$ & $10.83$& $7.66$\textsc{E-05} & $14.15$ & $3.84\textsc{E-04}$ & $\mathbf{10.32}$ & $2.35\textsc{E-06}$ \\
        \midrule
        Unicycle (sma.) & $606.11$ & $-$ & $157.92$&$5.88$\textsc{E-03} & $136.52$ & $\mathbf{1.94}\textsc{E-02}$ & $\mathbf{50.18}$ & $1.67\textsc{E-02}$ \\
        Unicycle (med.) &$1524.04$ & $-$ & $126.33$& $5.89\textsc{E-03}$ &$116.70$ &$\mathbf{3.43}\textsc{E-02}$ & $\mathbf{36.41}$ & $1.47\textsc{E-02}$\\
        Unicycle (lar.) & $1001.18$ & $-$&$219.07$ & $5.85\textsc{E-03}$&$190.62$& $8.59\textsc{E-03}$ & $\mathbf{73.17}$ & $\mathbf{1.61}\textsc{E-02}$\\
        \midrule
        Attitude (sma.) & $557.85$ & $-$ & $18040.96$&$1.31\textsc{E-04}$ & $\mathbf{5297.40}$ & $\mathbf{6.86}\textsc{E-04}$ & $13828.22$ & $4.05\textsc{E-04}$ \\
        Attitude (med.) &$\times$ & $\times$ & $18528.36$& $1.31\textsc{E-04}$ &$\mathbf{5724.30}$ &$\mathbf{6.68}\textsc{E-04}$ & $14279.95$ & $4.47\textsc{E-04}$\\
        Attitude (lar.) & $\times$ & $\times$&$17834.08$ & $1.31\textsc{E-04}$&$\mathbf{5400.88}$& $\mathbf{6.83}\textsc{E-04}$ & $13454.96$ & $4.05\textsc{E-04}$\\
        \bottomrule
    \end{tabular}}
    \setlength{\abovecaptionskip}{10pt}\caption{Benchmark computation time (s) and set volumes. \ding{72} denotes our polyhedral-enclosure based FITS algorithm to solve Eq. \ref{eq:bw_Reach_inner}, $-$ denotes instances where the procedure runs but fails to return a set with nonzero volume, and $\times$ denotes a failure for any other reason}
    \label{tab:results-transposed2}
\end{table*}
\vspace*{-4em}
\subsection{Combining Forward and Backward Analysis}
Finally, we report results for combining forward and backward reachability analysis. We combine the forward analysis method of~\cite{akinwande2025verifying} with the refined \dripy algorithm and with the \textsc{CRISP} variant of the \textsc{FITS} algorithm. We compare these results against simply using the forward algorithm of~\cite{akinwande2025verifying} alone.

A single run of forward analysis suffices to verify the forward reach--avoid properties (eqs.~\eqref{eq:fwd_reach_prop} and~\eqref{eq:fwd_avoid_prop}). In contrast, verifying the backward reach--avoid properties (eqs.~\eqref{eq:bw_reach_prop} and~\eqref{eq:bw_avoid_prop}) requires two separate backward-analysis runs, corresponding to the computation of the must-BRS and the may-BRS. We report these results separately.

On relatively easy benchmarks such as TORA, the overhead introduced by backward analysis dominates the overall computational cost, rendering the \fab strategy detrimental. As the benchmarks become more challenging, however, the \fab strategy becomes beneficial. On the Unicycle benchmark, we observe speedups of up to $7\times$ for both reach and avoid properties. Even after accounting for the fact that forward-only verification requires only a single forward-analysis run, the \fab strategy still improves over forward-only analysis by a factor of $2$--$3\times$. The results are mixed on the Attitude benchmark, where we observe improvements of up to $1.7\times$ on the avoid specification, but the \fab strategy appears detrimental for the reach specification. When we account for the need for a single forward-only analysis run, \fab requires up to $2\times$ the amount of time needed to verify the same set of properties. 

\begin{table*}[h!]
    \centering
    \begin{tabular}{lccccc}
        \toprule
        & Forward analysis
        & \multicolumn{2}{c}{\textsc{FaBRIC}(\ding{72})}
        & \multicolumn{2}{c}{Improvement} \\
        \cmidrule(lr){3-4}
        \cmidrule(lr){5-6}
        & 
        & Reach & Avoid
        & Reach ($\downarrow$) & Avoid ($\downarrow$) \\
        \midrule
        Tora (Small)   & $1.99$ & $10.87$ & $6.41$ & $-5.46$ & $-3.22$ \\
        Tora (Med)     & $2.21$ & $10.09$ & $6.61$ & $-4.57$ & $-2.99$ \\
        Tora (Large)   & $3.29$ & $16.70$ & $8.92$ & $-5.08$ & $-2.71$ \\
        \midrule
        Unicycle (Small)  & $1302.98$ & $280.51$ & $292.02$ & $4.65$ & $4.46$ \\
        Unicycle (Medium) & $1157.98$ & $261.27$ & $260.15$ & $4.43$ & $4.45$ \\
        Unicycle (Large)  & $2928.21$ & $384.56$ & $408.82$ & $7.61$ & $7.16$ \\
        \midrule
        Attitude (Small)  & $6980.36$ & $8365.28$ & $4001.11$ & $-1.20$ & $1.75$ \\
        Attitude (Medium) & $5823.11$ & $7349.67$ & $3458.57$ & $-1.26$ & $1.68$ \\
        Attitude (Large)  & $7223.28$ & $10658.96$ & $4182.43$ & $-1.48$ & $1.73$ \\
        \bottomrule
    \end{tabular}
    \setlength{\abovecaptionskip}{10pt}
    \caption{Comparison of \fab and Forward-only analysis with measured improvements}
    \label{tab:forward-only-fabric-improvement}
\end{table*}

\section{Conclusions}
We introduced \fab, a new strategy for verifying reach--avoid specifications in nonlinear neural feedback systems. The \fab strategy is enabled by a set of new algorithms that soundly approximate may- and must-backward reachable sets for nonlinear neural feedback systems. Using a representative benchmark suite, we demonstrated that our backward reachable set approximation algorithms significantly improve on the prior state of the art. Finally, we showed that the \fab strategy can substantially reduce the computational effort required to verify reach--avoid specifications in neural feedback systems.
\paragraph{Future Work}
We plan to explore further optimizations for inner-set estimation, as the current methods remain sensitive to the curse of dimensionality. One promising direction is to use sampling to refine the domain prior to running the \textsc{FITS} algorithm. We also intend to investigate the meta-optimization problem of selecting $F$ and $B$, with the goal of automating this selection process.

\newpage
\bibliographystyle{plain}
\bibliography{refs}
\newpage
\appendix
\section{Overapproximating the Forward Reachable Set}
For completeness, we describe an approach to computing a overapproximate forward query.
We can compute sound forward sets by solving Equation~\ref{eq:fwd_diffEq} over $\horizon$ time steps for all states in $\stateBig_0$. 
We employ the method of \cite{akinwande2025verifying} to compute an overapproximation of the true forward reachable set.
We can abstract the nonlinear transition functions in Equation~\ref{eq:fwd_diffEq} by computing tight polyhedral enclosures (upper and lower bounds) for each function. 
These enclosures, together with the neural network controller $\ctrl$, can be encoded as mixed-integer linear programs (MILPs), which enable us to overapproximate the forward reachable set $\nextt^{\dynsys}(\stateBig_0)$.

Specifically, we can compute a lower bound on $\nextt^{\dynsys}(\stateBig_0)$ by solving the following optimization problem for each $i \in [1, \ldots, n]$:
\begin{subequations}
\begin{align}
    \min_{\state \in \hat{\stateBig}} \; \state_i + \big(\vec{y}_i + \ctrl(\state)_i + \epsilon_i\big) \cdot \timestep, \\
    \text{s.t.} \quad \quad  \model_0, \dots, \model_n, \quad \epsilon \in \noise,
\end{align}
\label{eq:fw_Reach}
\end{subequations}
where $\model_0, \dots, \model_n$ denote the MILP constraints representing the system dynamics and controller. 
Replacing the minimization with a maximization yields an overapproximation of the upper bound. 
The resulting hyperrectangle ($\stateBig_1$) is guaranteed to enclose the true forward reachable set $\nextt^{\dynsys}(\stateBig_0)$. Sequential application of this process using $\stateBig_i$ for $i \in [0..T-1]$ is called \emph{concrete} reachability analysis \cite{sidrane2022overt}. 

Note that this process approximates arbitrary sets using hyperrectangles. Sequential application of \Cref{eq:fw_Reach} may lead to a coarse overapproximation of the forward reachable set, a situation referred to as \emph{excess conservatism}.
To address this, we can instead represent two steps at once.  For $i\in[0..n]$, let $\model'_i$ be the same as $\model_i$, except that each variable $v$ appearing in $\model_i$ is replaced by $v'$ in $\model'_i$.  To solve the two-step problem, we then solve:
\begin{subequations}
\begin{align}
    \min_{\state \in \hat{\stateBig}_t} \state'_i + (y'_i + \ctrl'(\state')_i + \epsilon') \cdot \delta \\
    \quad \model_0, \dots, \model_n\\
    \text{for each }i\in[n]
\begin{cases}
\state'_i = \state_i + (y_i + \ctrl(\state)_i + \epsilon_i),\quad \epsilon_i\in\noise
\end{cases}\\
        \quad \model'_0, \dots, \model'_n \quad \epsilon' \in \noise
    \label{eq:SymReach}
\end{align}
\noindent
This yields lower bounds for $\state_{t+2}$ (as before, replacing $\min$ by $\max$ produces upper bounds).
This process can be repeated to generalize from a two-step analysis to a $k$-step analysis for arbitrary $k$. Constructing $\hat{\ftraj}^\dynsys(I)$ this way is called \emph{symbolic} reachability analysis \cite{sidrane2022overt}.
\end{subequations}
\section{Benchmarks}
\subsubsection{TORA}
We would like to verify that the states of an actuated cart remain within a safe region during a specified time interval. Formally, we define the neural feedback system $\tora$, where $n^\tora = 4$, 
\begin{align*}
    &\dom^\tora = [-6.0,6.0] \times [-6.0, 6.0] \times [-3.14,3,14] \times [-2.0,2.0] \\
    &\init^\tora = [0.6, 0.7] \times [-0.7, -0.6] \times [-0.4,-0.3] \times [0.5, 0.6]
\end{align*}
the transition functions are defined as
\begin{align*}
    \trans^{\tora}_i = 
    \begin{cases}
       \vec{x}_2 &\text{if} \quad i = 1 \\
       -\vec{x}_1+0.1\sin(\vec{x}_3) &\text{if} \quad i = 2 \\
       \vec{x}_4 &\text{if} \quad i = 3 \\
        0 &\text{if} \quad i = 4 
    \end{cases}
\end{align*}
$u^\tora$ is computed by a neural network with three hidden layers (each with $k$ neurons), and four outputs, the first three of which is set to the constant zero value, $\delta^\tora = 0.1$, $T^\tora = 20$, 
\begin{align*}
    &G^\tora = [0.5,1.5] \times [-0.5,0.5] \times [-0.2,0.2] \times [-0.2,0.2] \\
    &A^\tora = [-0.2,0.2] \times [-0.2,0.2] \times [-0.2,0.2] \times [-0.2,0.2] \quad \textbf{for every} \quad t \in [T]
\end{align*}
We would like to verify that the system is safe for controllers with $k \in \{50,100,200\}$. \vspace{-1em}
\subsubsection{Unicycle Car Model}
We would like to verify that a simplified model of an autonomous vehicle is safe. The car is modeled with four state variables, representing the $ x $ and $ y $ coordinates in a plane, the steering angle, and the velocity magnitude.
Formally, we define the neural feedback system $\uni$, where $ n^{\uni}=4$, 
\begin{align*}
    &\dom ^\uni = [-12.0,12.0] \times [-12.0, 12.0] \times [-\pi,\pi] \times [-1,1] \\
    &\init^{\uni} = [9.5,9.55] \times [-4.5,-4.45] \times [2.1,2.11] \times [1.5,1.51]
\end{align*}
The transition functions are defined as s
\begin{align*}
    \trans^{\uni}_i = 
    \begin{cases}
       \state_4\cos(\state_3) &\text{if} \quad i = 1 \\
       \state_4\sin(\state_3) &\text{if} \quad i = 2 \\
        0 &\text{if} \quad i = 3 \\
        0 &\text{if} \quad i = 4 
    \end{cases}
\end{align*}
$ \noise = \{0\}\times\{0\}\times\{0\}\times [-10^{-4},10^{-4}]$, $ \ctrl^{\uni} $ 
is computed by a neural network with one hidden layer with $k$ neurons and four outputs, the first two of which are set to the constant zero value, $ \timestep^{\uni} = 0.2 $ , $ \horizon^{\uni} = 50 $, 
\begin{align*}
    &\reach^{\uni} =[1.5,2.5] \times [1.5,2.5] \times [-0.5,0.5] \times [-0.5,0.5] \\
    &\avoid^{\uni}(t) = [-0.2,0.2] \times [-0.2,0.2] \times [-\pi,\pi], [-2,2] \quad \textbf{for every} \quad t \in [T]
\end{align*}
We would like to verify that the system is safe for controllers with $k \in \{250,500,1000\}$
\vspace{-1em}
\subsubsection{Attitude Control Model}
We would like to verify that a simplified model of an autonomous aircraft is safe. The aircraft is modeled as a ridid body with six state variables, representing the orientation ($\psi_1,\psi_2,\psi_3$) using Rodrigues parameters, and rotation rates ($\omega_1,\omega_2,\omega_3$) of the aircraft.
Formally, we define the neural feedback system $\att$, where $ n^{\att}=6$, 
\begin{align*}
    &\dom ^\att = [-7.0,7.0] \times [-7.0,7.0] \times [-7.0,7.0] \times [-3,3] \times [-3,3] \times [-3,3] \\
    &\init^{\att} =[4.5,5.0] \times [3.2,3.7] \times [-2.5,-2.0] \times [-0.1,0.1] \times [-0.1,0.1] \times  [-0.1,0.1]
\end{align*}
Let $\Psi = \psi_1^2 + \psi_2^2 + \psi_3^2$, then the transition functions are defined as
\begin{align*}
    \trans^{\att}_i = 
    \begin{cases}
        0.5(\omega_2(\Psi - \psi_3) + \omega_3(\Psi + \psi_2) + \omega_1(\Psi +1)) &\text{if} \quad i = 1 \\
        0.5(\omega_1(\Psi + \psi_3) + \omega_3(\Psi - \psi_1) + \omega_2(\Psi +1)) &\text{if} \quad i = 2 \\
        0.5(\omega_1(\Psi - \psi_2) + \omega_2(\Psi + \psi_1) + \omega_3(\Psi +1)) &\text{if} \quad i = 3 \\
        0.25(\omega_2\omega_3) &\text{if} \quad i = 4 \\
        -1.5(\omega_1\omega_3) &\text{if} \quad i = 5 \\
        2(\omega_1\omega_2) &\text{if} \quad i = 6
    \end{cases}
\end{align*}
$ \noise = \{0\}\times\{0\}\times\{0\}\times\{0\}\times\{0\}\times\{0\}$, $ \ctrl^{\att} $ 
is computed by a neural network with three hidden layers, (each with $k$ neurons) and six outputs, the first three of which are set to the constant zero value, $ \timestep^{\att} = 0.1$ , $ \horizon^{\att} = 10 $, 
\begin{align*}
    &\reach^{\att} = [-0.5,0.5] \times [-0.5,0.5] \times [-0.5,0.5] \times [-0.4,0.6] \times [-0.5,0.5] \\  &\hspace{3em}\times [-0.6,0.4] \\
    &\avoid^{\att}(t) = [-0.75,-0.55] \times [0.65,0.85] \times [-0.45,0.15] \times [-0.25,0.05] \\ & \hspace{3em} \times [-0.55,-0.35] \times [-0.05,0.25]
    \quad \textbf{for every} \quad t \in [T]
\end{align*}
We would like to verify that the system is safe for controllers with $k \in \{32,64,128\}$

\end{document}